\documentclass[11pt,a4paper]{article}
\usepackage{times,latexsym}
\usepackage{url}
\usepackage[T1]{fontenc}
\usepackage{hhline}
\usepackage{hyperref}
\usepackage{float}
\usepackage{array}
\usepackage{multirow}
\usepackage{placeins}
\usepackage{makecell}
\usepackage{enumitem}

\usepackage[acceptedWithA]{tacl2021v1}
\usepackage{amsmath}

\usepackage{xspace,mfirstuc,tabulary}

\usepackage{graphicx}
\usepackage{tabularx}

\newcommand{\sref}[1]{\S\ref{#1}} %
\newcommand{\genadv}{\textbf{Gen}erative model-based \textbf{ADV}ersarial attack (\textbf{GenADV})}

\newcommand{\RAD}{\textbf{R}obustness under \textbf{A}dditional \textbf{D}ocument (\textbf{RAD})}
\newcommand{\ARA}{\textbf{A}ccurate set of \textbf{R}etrieval \textbf{A}ugmentation}
\newcommand{\MainSubject}{imperfect documents}

\definecolor{RoseQuartzBg}{HTML}{F7CAC9}
\definecolor{RoseQuartz}{HTML}{F5A798}
\definecolor{Serenity}{HTML}{92A8D1}
\definecolor{OrangeRed}{rgb}{1.0, 0.27, 0.0}
\definecolor{Red}{rgb}{1.0, 0.0, 0.0}
\definecolor{Turquoise}{HTML}{0F4C81}
\usepackage{xparse}
\NewDocumentCommand{\nishant}{ mO{} }{\textcolor{blue}{\textsuperscript{\textit{Nishant}}\textsf{\textbf{\small[#1]}}}}
\NewDocumentCommand{\wenlong}{ mO{} }{\textcolor{Serenity}{\textsuperscript{\textit{Wenlong}}\textsf{\textbf{\small[#1]}}}}
\NewDocumentCommand{\jy}{ mO{} }{\textcolor{RoseQuartz}{\textsuperscript{\textit{jy}}\textsf{\textbf{\small[#1]}}}}
\NewDocumentCommand{\jyc}{ mO{} }{\textcolor{blue}{\textsuperscript{\textit{jyc}}\textsf{\textbf{\small[#1]}}}}
\NewDocumentCommand{\js}{ mO{} }{\textcolor{OrangeRed}{\textsuperscript{\textit{jong}}\textsf{\textbf{\small[#1]}}}}
\NewDocumentCommand{\jsc}{ mO{} }{\textcolor{blue}{\textsuperscript{\textit{jong comment}}\textsf{\textbf{\small[#1]}}}}

\renewcommand{\nishant}[2][]{}
\renewcommand{\wenlong}[2][]{}

\newif\iftaclinstructions
\taclinstructionsfalse 
\iftaclinstructions
\renewcommand{\confidential}{}
\renewcommand{\anonsubtext}{(No author info supplied here, for consistency with
TACL-submission anonymization requirements)}
\newcommand{\instr}
\fi

\iftaclpubformat 

\else

\fi

\title{Toward Robust RALMs: Revealing the Impact of Imperfect Retrieval on Retrieval-Augmented Language Models}

\author{
  Seong-Il Park 
  \\
  Graduate School of Data Science, 
  \\
  Seoul National University
  \\
  Seoul, Republic of Korea
  \\
  \texttt{athjk3@snu.ac.kr}
  \And
  Jay-Yoon Lee
  \\
  Graduate School of Data Science, 
  \\
  Seoul National University
  \\
  Seoul, Republic of Korea
  \\
  \texttt{lee.jayyoon@snu.ac.kr}
}

\usepackage{array}
\usepackage{etoolbox}

\AtBeginEnvironment{tabular}{\fontsize{9.5}{13.5}\selectfont}
\begin{document}
\maketitle
\begin{abstract}
    Retrieval Augmented Language Models (RALMs) have gained significant attention for their ability to generate accurate answer and improve efficiency. 
    However, RALMs are inherently vulnerable to imperfect information due to their reliance on the imperfect retriever or knowledge source. 
    We identify three common scenarios-unanswerable, adversarial, conflicting-where retrieved document sets can confuse RALM with plausible real-world examples.
    We present the first comprehensive investigation to assess how well RALMs detect and handle such problematic scenarios. 
    Among these scenarios, to systematically examine adversarial robustness we propose a new adversarial attack method, \genadv\ 
    and a novel metric \RAD. 
    Our findings reveal that RALMs often fail to identify the unanswerability or contradiction of a document set, which frequently leads to hallucinations.
    Moreover, we show the addition of an adversary significantly degrades RALM’s performance, with the model becoming even more vulnerable when the two scenarios overlap (adversarial+unanswerable). 
    Our research identifies critical areas for assessing and enhancing the robustness of RALMs, laying the foundation for the development of more robust models.\footnote{The code and data can be found at \url{https://github.com/Atipico1/robust-rag}}
\end{abstract}


\section{Introduction}
Large Language Models (LLMs) are becoming the foundation for various NLP tasks \citep{brown2020language, anil2023palm, achiam2023gpt, qin2023chatgpt}. Notably, in open-domain question answering (QA) tasks \citep{chen2017reading} that require substantial knowledge, Retrieval Augmented Language Models (RALMs) have proven to be highly effective \citep{lewis2020retrieval,guu2020retrieval,izacard2021leveraging,izacard2022few,lin2023ra}. RALMs generate answers based on external knowledge and shows competitive performance with simple in-context setting without additional training. \citep{ram2023context}

\begin{figure}[ht]
    \centering
    \includegraphics[width=0.95\columnwidth]{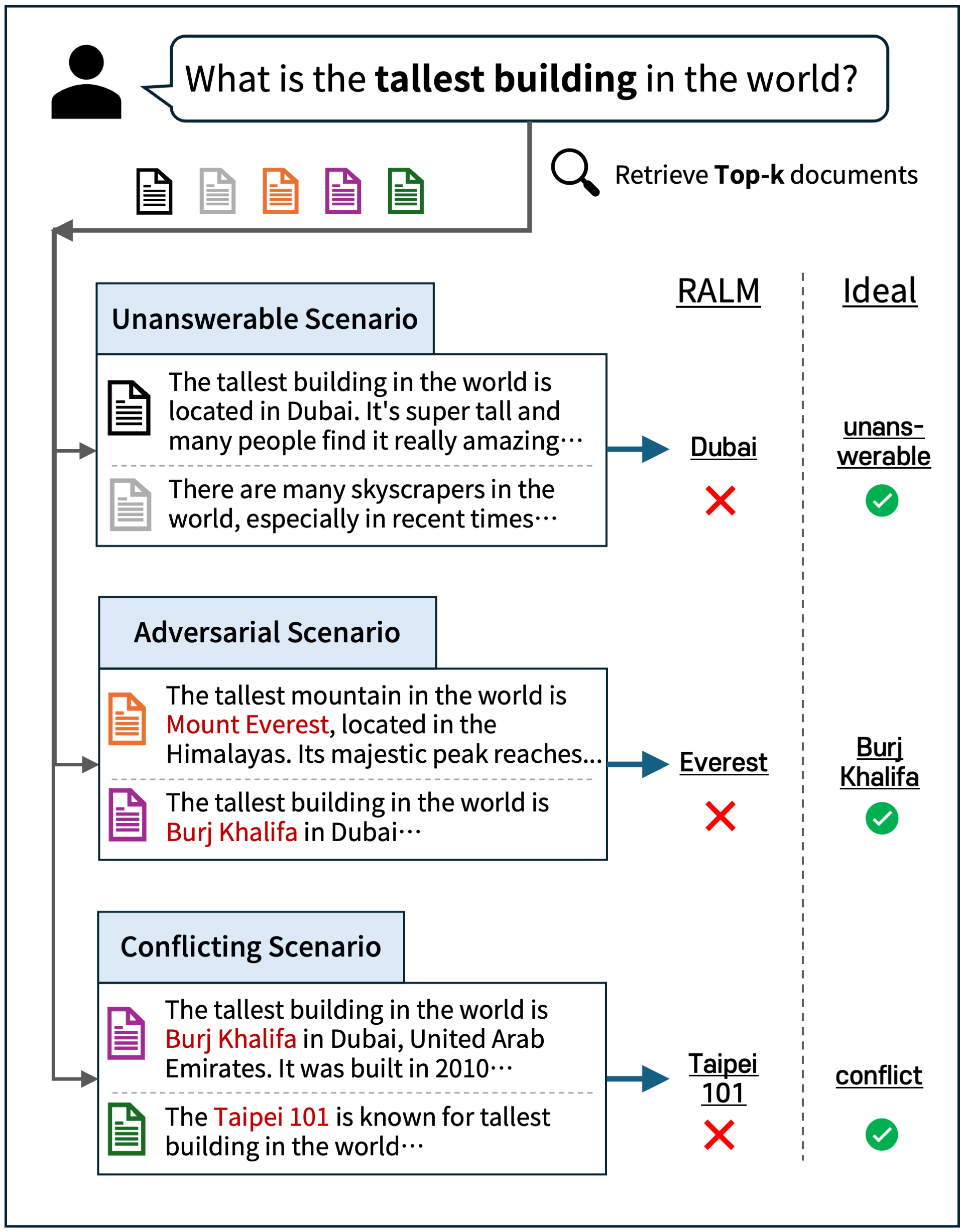}
    \caption{Examples of scenarios with imperfect information. A Robust RALM system can be resilient to imperfections inherent in search engines or knowledge sources.}
    \label{fig:intro}
\end{figure}

However, RALMs are known to be sensitive to the quality of external information due to their reliance on it. 
Common issues such as imperfect retriever or contaminated knowledge sources can affect the robustness of RALMs. \citep{petroni2020context, du2022synthetic, li2023large, du2022synthetic}
For example, Figure \ref{fig:intro} shows different types of real-world scenarios, illustrating both incorrect responses by RALM and their ideal responses. 
If a query "What is the tallest building in the world?" retrieves documents that do not contain the answer, the RALM should classify it as "unanswerable" rather than parroting incorrect answer in the document (\textit{unanswerable scenario}). 
Furthermore, RALM should ignore documents that do not contain the answer even if it appears to be related to the question (e.g., "The tallest mountain in the world is Mount Everest") and instead extract the answer from documents that do (\textit{adversarial scenario}). 
In cases where documents provide conflicting answers (e.g., "The tallest building in the world is Burj Khalifa... The Taipei 101 is known for the tallest building in the world"), the model should respond with "conflict" as RALM cannot surely identify which is the correct answer (\textit{conflicting scenario}).

Previous studies have primarily focused on one of these three scenarios, \citep{chen2022rich, weller2022defending, ren2023investigating} or on inconsistencies within individual documents \citep{longpre2021entity}, addressed methods for mitigating \citep{asai2023self, yu2023chain, xu2023recomp} or examined the relationship between parametric knowledge and documents rather than interactions across multiple retrieved documents \citep{xie2023adaptive}. 
In contrast, our work systematically analyzes the robustness of RALMs for open-domain QA in scenarios of imperfect information which is a critical factor of RALM's robustness. 
We define each type of scenario and develop a perturbation method to generate imperfect documents, particularly for simulating adversarial and conflict scenarios in open-domain QA.
We also introduce appropriate robustness metrics for each experiment. 



For unanswerable scenario, we categorize examples into answerable and unanswerable based on whether the answer strings appear in the retrieved documents, and then measure the accuracy for each subset. 
Our findings show the challenges RALMs encounter in accurately identifying unanswerability. 
This often leads to hallucination or parroting incorrect answer in the documents instead of saying unanswerable.
For adversarial scenario (Table \ref{tab:ex_adv}), we introduce a new adversarial attack framework, \genadv\, and propose a new metric, \RAD\, to measure adversarial robustness of RALMs. 
Our results indicate that RALMs are vulnerable to adversarial information, particularly GenADV. 
Also, they show higher vulnerability in adversarial-unanswerable situations where an unanswerable example contain adversarial information in the retrieved documents.
Concerning conflicting documents (Table \ref{tab:ex_conflict}), we investigate how well RALMs detect the conflicts and base their responses on conflicting information.
This highlight the difficulties RALMs face in detecting conflicts and how easily they can be misled by such misinformation.

In summary, our main contributions are as follows:
\begin{itemize}[noitemsep]
    \item We identify common scenarios involving imperfect information that frequently occur in real-world retrievers, and develop perturbation techniques to simulate these scenarios.
    \item We design experiments to assess robustness in scenarios of imperfect information and proposed corresponding metrics for a systematic analysis of the results.
    \item We propose a new adversarial attack method, GenADV, and a metric, RAD, specifically designed to measure adversarial robustness in open-domain QA systems.
    \item We conduct experiments to evaluate how effectively RALMs detect imperfect information and how often they hallucinate in such situations.
\end{itemize}
\section{Related Works}
\textbf{In-context RALMs} Traditionally, Retrieval-Augmented Language Models (RALMs) involved training a generator to generate answers based on the retrieved documents \citep{lewis2020retrieval, izacard2021leveraging, izacard2022few}. However, recent discoveries show that LLMs can be used as generators for RALM in an in-context setting, without additional training, by simply concatenating retrieved documents to the query \citep{levine2022huge, levine2022standing, kamalloo2023evaluating, shi2023replug,ram2023context}. 
Since this method is highly efficient and promising in open-domain QA, we will use the in-context RALMs.

\textbf{Robustness of LLMs on imperfect information} Recent work has demonstrated that LLMs are sensitive to imperfect information, revealing a tendency to adhere to parametric knowledge acquired during pre-training when it conflicts with the given context \citep{longpre2021entity, chen2022rich, xie2023adaptive}.
In contrast, our research aims to determine whether LLMs can accurately identify conflicts among multiple documents in a retrieval scenario and ascertain the basis for their responses. 
Additionally, other previous works have shown that LLMs can produce incorrect answers, a phenomenon often referred to as 'hallucination', especially when the available information is insufficient \citep{asai2021challenges, ren2022out, sulem2022yes, hu2023won, ren2023investigating}. 
Building on these findings, our study shifts focus to assess how well LLMs can identify unanswerability in complex scenarios, providing a thorough analysis of situations where detection fails.
Moreover, various studies have shown that LLMs can be easily distracted by irrelevant information \citep{jia2017adversarial,  petroni2020context, creswell2022selection, cao-etal-2022-tasa, shi2023large, yoran2023making}. To expand on these findings, we demonstrate how to generate distracting information in open-domain QA scenarios to assess the robustness of RALMs.


\section{Problem Setup}
\subsection{Definition of RALM} Our RALM follows in-context RALM framework \citep{ram2023context}, with a particular focus on open-domain QA.

In in-context RALM, for given a input query $q$ and generated response $y$, we retrieve documents from external knowledge source and use the $k$ highest ranked documents $d = [d_1, d_2, …, d_k]$. We then concatenate $q$ with $d$ for generation. The generation process is represented as:
\begin{equation}
    {p(y|q)} = {p(y|d, q)p(d|q)}
\end{equation}

LLMs can directly generate answers for open-domain QA directly through prompting \citep{levine2022standing, levine2022huge}. Therefore, we utilize a frozen LLM as the generator in our RALM.

\subsection{Types of \MainSubject}
Our experiments address three scenarios of imperfect information in open-domain QA. Each represents a scenario frequently encountered in retrieval for open-domain QA.

\textbf{Unanswerable Scenario} The first scenario involves unanswerability, in which the set of retrieved documents lacks sufficient information to answer the provided query.
In this scenario, there is a high likelihood of hallucination which means parroting the incorrect answer in the documents when RALMs generate responses, thus abstaining is crucial. In our experiments, an unanswerability is identified when all the top-k retrieved documents do not contain the answer string. For detailed experimental settings, refer to section \ref{subsec:unans_setting}.

\textbf{Adversarial Scenario}  The second scenario, adversarial information refers to situations where the correct answer is not included in the retrieved document, yet the model is misled by distracting information in that document. 
Table \ref{tab:ex_adv} displays real examples of adversarial information present in the open-domain QA dataset, indicating that RALMs can easily be distracted by such adversarial information.
Unlike prior studies that primarily focus on adversarial attacks in Machine Reading Comprehension (MRC) systems \citep{jia2017adversarial, cao-etal-2022-tasa} or classification task \citep{pruthi2019combating, li2021contextualized, lei2022phrase}, our study addresses adversarial attacks in the context of open-domain QA which utilizes multiple documents for generation.


\begin{table}[H]
\centering
\resizebox{\columnwidth}{!}{%
\begin{tabular}{cl}
\hline
\multicolumn{2}{c}{\textbf{TQA}} \\ \hline
\multicolumn{1}{c|}{Q} & {What is the largest city in Ohio?} \\ \hline
\multicolumn{1}{c|}{A} & Cleveland  \ \ \ \textcolor{red}{(Cincinnati)} \\ \hline
\multicolumn{1}{c|}{\begin{tabular}[c]{@{}c@{}} Docs\end{tabular}} & \begin{tabular}[c]{@{}l@{}}{[Doc1]} \textbf{Cincinnati} is the \textbf{third-largest} city in Ohio and \\ 65th in the United States. Its metropolitan area is ... \\ {[Doc2]} This makes \textbf{Dayton} the \textbf{fourth-largest} metropol-\\ itan area in Ohio and 63rd in the United States... \end{tabular} \\ \hline
\multicolumn{2}{c|}{\textbf{NQ}} \\ \hline
\multicolumn{1}{c|}{Q} & Who got the first nobel prize in physics? \\ \hline
\multicolumn{1}{c|}{A} & Wilhelm Conrad Röntgen \ \ \ \textcolor{red}{(Yuval Katzenelson)} \\ \hline
\multicolumn{1}{c|}{\begin{tabular}[c]{@{}c@{}}Docs\end{tabular}} & \begin{tabular}[c]{@{}l@{}}$[$Doc1$]$ In 2012, the \textbf{first prize winner} was another Israeli \\ teenager, \textbf{Yuval Katzenelson} of Kiryat Gat, who presented...\\ {[Doc2]} ... three names on the list: \textbf{Werner Heisenberg},\\ who received the \textbf{Nobel Prize in Physics} in 1932...\end{tabular} \\ \hline
\end{tabular}%
}
\caption{Actual examples included in the dataset (TriviaQA, Natural Questions) and retrieved documents (Docs) containing adversarial information. \textcolor{red}{Red text} indicates the actual output of RALM.}
\label{tab:ex_adv}
\end{table}

\begin{table}[]
\centering
\resizebox{\columnwidth}{!}{%
\begin{tabular}{cl}
\hline
\multicolumn{2}{c}{\textbf{TQA}} \\ \hline
\multicolumn{1}{c|}{Q} & {What is a “Scotch Bonnet”?} \\ \hline
\multicolumn{1}{c|}{A} & Chili Pepper  \ \ \ \textcolor{red}{(Sea snail)}  \\ \hline
\multicolumn{1}{c|}{\begin{tabular}[c]{@{}c@{}} Docs\end{tabular}} & \begin{tabular}[c]{@{}l@{}}{[Doc1]} Scotch bonnet (Semicassis granulata) is a medium- \\ sized to large species of \textbf{sea snail}, a marine gastropod... \\ {[Doc2]} Scotch bonnet, also known as bonney peppers,\\ or Caribbean red peppers, is a variety of \textbf{chili pepper}...\end{tabular} \\ \hline
\multicolumn{2}{c|}{\textbf{NQ}} \\ \hline
\multicolumn{1}{c|}{Q} & How many countries are a part of opec? \\ \hline
\multicolumn{1}{c|}{A} & 14  \ \ \ \textcolor{red}{(15)}  \\ \hline
\multicolumn{1}{c|}{\begin{tabular}[c]{@{}c@{}}Docs\end{tabular}} & \begin{tabular}[c]{@{}l@{}}{[Doc1]} As of June 2018, OPEC has \textbf{15 member countries:}\\ six in the Middle East (Western Asia), seven in Africa, ... \\ {[Doc2]} As of May 2017, OPEC consists of \textbf{14} \textbf{countries}\\ which earn the majority of their income...\end{tabular} \\ \hline
\end{tabular}%
}
\caption{Actual examples included in the dataset (TriviaQA, Natural Questions) and retrieved documents (Docs) containing conflicting information. \textcolor{red}{Red text} indicates the actual output of RALM.}
\label{tab:ex_conflict}
\end{table}

\textbf{Conflict Scenario}  The last scenario deals with conflicting information, where there is a contradiction among the information in the retrieved documents. 
Table \ref{tab:ex_conflict} shows real example of conflicting information in the dataset. 
This is a common situation that can occur especially with search engine results, due to the multiple sources involved, or with information that changes over time. 
Additionally, intentional information poisoning can contaminate knowledge sources \citep{du2022synthetic, pan2023attacking}, making it crucial to detect and resolve conflicting information. 
In our experiments, we follow similar strategies to those described in \citet{xie2023adaptive} for creating conflicting information, in order to test RALM's ability to detect the conflict.
For detailed experimental settings, refer to \sref{subsubsec:craft_conflict}.

\section{Experimental Setup}
\label{sec:setup}
\subsection{Task and Datasets}
\label{subsec:task_and_dataset}
We conducted our experiments focusing on the open-domain QA. We utilized four benchmark datasets: Natural Questions (NQ) \citep{lee2019latent}, TriviaQA (TQA) \citep{joshi2017triviaqa}, Web Questions (WebQ) \citep{berant2013semantic}, and PopQA \citep{mallen2023not}.
We retrieved the top five documents for each question from Wikipedia\footnote{We used preprocessed data following \citep{karpukhin2020dense}} based on their cosine similarity to the questions and generated answers using these documents. All our experiments, except for PopQA, were performed on test sets. For the details, please refer to the Table \ref{tab:dataset}.

\subsection{Metrics} We use accuracy \citep{mallen2023not} as a primary metric. Unlike Exact match score, we consider a prediction correct if any substring of the prediction exactly matches any of the answers. This emphasis aligns with our goal to test the robustness of models rather than their extractive QA capabilities from 
LLMs. Specific metrics defined to test robustness in each experiment will be discussed in the respective experimental sections.

\begin{table}[ht]
\centering
\resizebox{0.75\columnwidth}{!}{%
\begin{tabular}{l|ccc}
\hline
\multicolumn{1}{c|}{Datasets} & Size & Recall@5 & Unans \\ \hline
NQ & 3610 & 0.68 & 0.32 \\
TQA & 11313 & 0.76 & 0.24 \\
WebQ & 2032 & 0.65 & 0.35 \\
PopQA & 14267 & 0.67 & 0.33 \\ \hline
\end{tabular}%
}
\caption{Dataset statistics and Recall for Top-5 retrieval. Recall means top-k retrieval accuracy as used in \citep{karpukhin2020dense}. Unans denotes the proportion of examples for which none of the top-5 documents contain the answer string.}
\label{tab:dataset}
\end{table}

\subsection{Models} We use ColBERTv2 \citep{santhanam2022colbertv2} as a dense retriever. 
We experimented following DPR style passage retrieval \citep{karpukhin2020dense}. 
The LLMs used for generating answers were publicly available instruction following models capable of RALM while being in a frozen state. 
Models included Llama2 chat \citep{touvron2023llama}, Mistral Instruct-v2 \citep{jiang2023mistral}, Orca2 \citep{mitra2023orca}, Qwen 1.5 chat \citep{bai2023qwen}, and Gemma instruct \citep{team2024gemma}. 
Our experiments were conducted using 7B size models, with additional analysis on larger sizes within the same family.
Additionally, we used OpenAI's \texttt{gpt-4o-mini-2024-07-18} API \footnote{For detailed information on the model, refer to \url{https://platform.openai.com/docs/models/gpt-4o-mini}.} (abbreviated as \textbf{GPT4o-mini}) as a closed-source model for further comparison.
To minimize randomness in the generative model, greedy decoding was used during generation, and all random seeds were fixed. 
For faster inference, we used vLLM \citep{kwon2023efficient} in all experiments.

\subsection{Prompting}
We crafted instructions to assess how well LLMs can detect unanswerability and conflicts in a zero-shot RALM setting. The primary focus is on enhancing a standard retrieval-augmented QA system by integrating capabilities to recognize unanswerability and identify conflicts within the provided documents. The types of prompts are as follows:
\begin{figure*}[]
    \centering
    \includegraphics[width=1.0\textwidth]{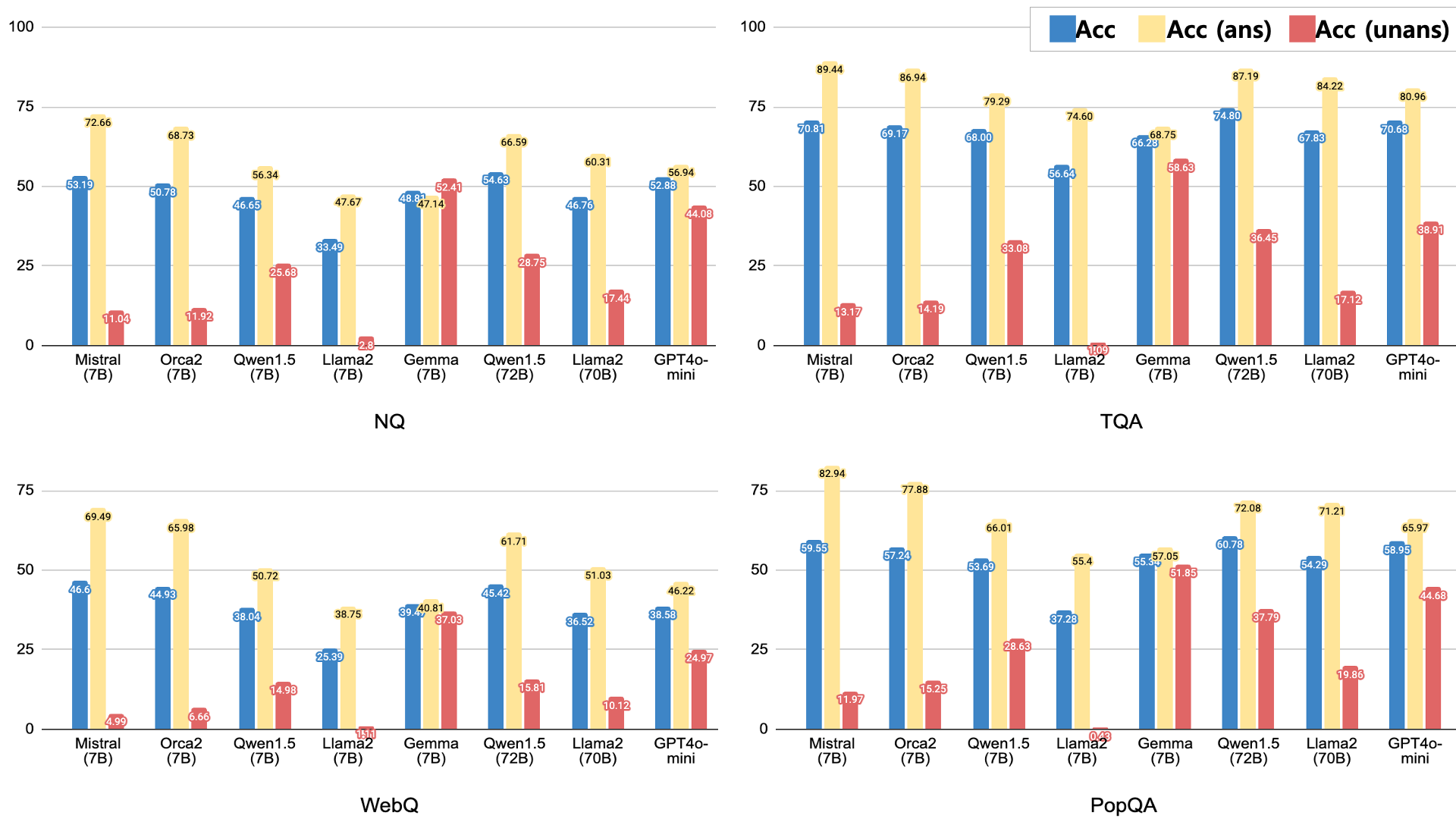}
    \caption{Experimental results on identifying unanswerable examples. The x axis represents the models (size). Acc means accuracy for all examples, Acc (ans) means accuracy for answerable examples and Acc (unans) means accuracy for unanswerable examples. The two models on the far right represent the results of experiments conducted on the largest models within their respective families.}
    \label{fig:unans_result}
\end{figure*}

\textbf{Normal prompt} This is our basic instruction for retrieval-augmented QA, enabling the LLM to utilize external information retrieved in response to questions.

\textbf{Unans prompt} This instruction incorporates unanswerability detection into the Normal prompt, requiring the LLM to not only search for answers but also assess whether the question can be answered based on the provided documents.

\textbf{Conflict prompt} This instruction introduces conflict detection to the Normal prompt, compelling the LLM to meticulously examine the retrieved data for any inconsistencies or contradictory information.

For the details of prompts, refer to Appendix \ref{sec:prompts}

\section{Experiments}
\subsection{Baseline QA Performance as a reference}
As a reference, we report standard QA performances on RALM and closed-book settings, without our curated prompts for the analysis, on the datasets we utilize in the main experiment.
\subsection{Identifying Unanswerability}\label{subsec:unans_setting}
In this experiment, we aim to test the zero-shot capability of the RALM system to detect when retrieved documents do not contain the answer to a question, known as selective prediction.
We will also test how much the models hallucinate in such situations. Since all our datasets are based on extractive QA, we determined the unanswerability of a question by checking if none of the top-5 retrieved documents contained the answer string (we refer to these as \textit{unanswerable examples}). Unlike \citep{ren2023investigating}, which studied the RALM's ability to determine unanswerability through an additional verification step, we directly test whether the model can identify unanswerability by changing the original answer to \textit{unanswerable}.
This is because, in the real world, directly identifying unanswerability allows us to choose other options, such as using a closed book method instead of RALM, or attempting retrieval again. Therefore, if no document included the answer string, we change the original answer to \textit{unanswerable}\footnote{We use the specific term \textit{unanswerable} instead of a more general expression (e.g., “I don’t know”) because LLMs showed better identification performance with \textit{unanswerable}.} and if the model responds with \textit{"unanswerable"}, it is considered accurate.
Additionally, we instructed the LLM with the \textit{Unans prompt} to indicate 
unanswerability when it cannot find an answer in the given documents.
\subsubsection{Results and Analysis}
\textbf{Answerable vs. Unanswerable} We assessed the zero-shot capability of RALMs by dividing test examples into answerable and unanswerable examples and calculating accuracy for each subset. 
Figure \ref{fig:unans_result} displays the results, which show significantly lower accuracy for unanswerable examples across most models and datasets. 
These results indicate that RALMs generally struggle to identify unanswerable scenarios, even in large models and commercial model with strong reasoning capabilities.
There were variations in performance among models; for instance, the Llama2 exhibited near-zero accuracy, whereas the Gemma demonstrated higher unanswerable accuracy on the NQ dataset. 
However, high unanswerable accuracy isn't always reliable. We examined how often models incorrectly responded \textit{"unanswerable"} to answerable examples on the NQ dataset. Gemma, despite high unanswerable accuracy, did this for 28.59\% of answerable examples, while Llama2 only 0.7\%. This suggests high unanswerable accuracy could simply result from a high propensity of answering \textit{"unanswerable"}, rather than truly identifying unanswerabilities.

\begin{table}[h]
\resizebox{\columnwidth}{!}{%
\begin{tabular}{cccc|ccc}
 & \multicolumn{3}{c|}{NQ} & \multicolumn{3}{c}{TQA} \\ \cline{2-7} 
\textbf{Models} & \textbf{Acc.} & \textbf{Hallu.} & \textbf{Cor.} & \textbf{Acc.} & \textbf{Hallu.} & \textbf{Cor.} \\ \hline
\multicolumn{1}{c|}{Llama2} & 2.8 & 95.79 & 1.4 & 1.09 & 94.06 & 4.85 \\
\multicolumn{1}{c|}{Mistral} & 11.04 & 84.31 & 4.65 & 13.17 & 76.37 & 10.46 \\
\multicolumn{1}{c|}{Orca2} & 11.92 & 82.03 & 6.05 & 14.19 & 74.3 & 11.51 \\
\multicolumn{1}{c|}{Qwen1.5} & 25.68 & 73.09 & 1.23 & 33.08 & 63.34 & 3.58 \\
\multicolumn{1}{c|}{Gemma} & 52.41 & 47.5 & 0.09 & 58.63 & 39.96 & 1.41 \\
\multicolumn{1}{c|}{Qwen1.5*} & 28.75 & 67.74 & 3.51 & 36.45 & 53.38 & 10.17 \\
\multicolumn{1}{c|}{Llama2*} & 17.44 & 79.67 & 2.89 & 17.12 & 73.43 & 9.45 \\
\multicolumn{1}{c|}{GPT4o-mini} & 44.08 & 53.90 & 2.02 & 38.91 & 52.95 & 8.14 \\ \hline
\end{tabular}%
}
\caption{Detailed experimental results for unanswerable examples in the NQ and TriviaQA. 
Acc. indicates percentage of examples where the model accurately identified a question as \textit{unanswerable} (same as Acc (unans) in Figure \ref{fig:unans_result}).
Cor. means examples where the model provided the true answer to the question. Hallu. represents examples that are neither Acc. nor Cor. * indicates largest models within the family (70B for Llama2 and 72B for Qwen1.5, respectively); all others are 7B models.}
\label{tab:hallu}
\end{table}

\textbf{Not Responding "Unanswerable" Does Not Imply Correctness} Table \ref{tab:hallu} categorizes the results for the unanswerable examples three groups: those that accurately identified the question as \textit{unanswerable} (abbreviated as \textbf{Acc.}), those that provided the original answer to the question (abbreviated as \textbf{Cor.}), and those that produced a hallucinated response (abbreviated as \textbf{Hallu.}). 
In all cases, the hallucination ratio significantly outweighed the corrects. Notably, in the NQ dataset, Llama2 hallucinated  95.79\% of the time, and 94.06\% in the TriviaQA dataset. 
The large-sized models exhibited similar trends. 
The Qwen1.5 (with 72B parameters) provide the original answer in only 3.51\% of the examples and Llama2 (with 70B parameters) did so in just 2.89\% of cases.
This demonstrates that failing to correctly respond "unanswerable" does not mean the model has provided the original correct answer; rather, it indicates that the models are mostly \textbf{hallucinating}.

\textbf{Model Size and Robustness} Figure \ref{fig:unans_result} also show the results of the larger models. Across all four datasets, larger models exhibited higher accuracy for both answerable and unanswerable examples than their smaller model.
Specifically, the Llama2 model, except on the Web Questions, showed a greater performance gain for unanswerable than for answerable examples.
This suggests that more complex models possess superior reasoning abilities in more challenging scenarios.
However, despite this, the relative accuracy for unanswerable examples remains very low in models larger than 70B, indicating that relying solely on LLM responses to identify unanswerability could be potentially risky.

\subsection{Robustness on Adversarial Information}
In this experiment, our objective is to test the RALM's robustness in generating correct answers when adversarial documents designed to distract the model are included in the retrieved documents. For the test, we developed a new adversarial attack method for open-domain QA.
\subsubsection{Crafting adversarial information}
Traditional adversarial information generation in QA systems typically uses word substitution at the entity level, suitable for MRC tasks that rely on a given gold context. \citep{jia2017adversarial, jin2020bert, cao-etal-2022-tasa}. However, this approach is less suitable for open-domain QA, which requires multiple passages and does not provide a gold context in advance. Additionally, such adversarially crafted sentences are often grammatically or contextually inconsistent with other documents. To address these issues, we generated adversarial passages using a \textbf{Gen}erative model based \textbf{ADV}ersarial attack \textbf{(GenADV)}. GenADV is a hybrid approach. By replacing entities within sentences with similar entities, it retains semantic similarity to the original sentence, while also leveraging LLM to enhance consistency and naturalness in adversaries. However, unlike previous approaches, which depended on gold context \citep{cao-etal-2022-tasa} or human annotation \citep{jia2017adversarial}, it relies solely on question-answer (Q-A) pairs. The following describes the process of generating adversarial information using GenADV (Table \ref{tab:genadv_ex}).
\begin{enumerate}[noitemsep,partopsep=0.1pt]
\item \textbf{Creating an answer sentence and detecting entities:} Initially, we generated an answer sentence using a Q-A pair.
After detecting all named entities in the created answer sentence, sentences containing fewer than two entities were excluded. The reason is that when only one entity is detected, there is a high likelihood of conflict if substitution is performed.
\item \textbf{Generating Adversarial sentence and filtering:} Using an LLM, we substituted entities in the answer sentence with similar ones to create an adversarial sentence that retained similar meanings but differed in information from the original answer sentence.
We excluded adversarial sentences that contained information about the correct answers. Specifically, we removed any sentence that included the answer string or exhibited a cosine similarity of 0.8 or higher with the answer sentence.\footnote{We used the Sentence Transformers \citep{reimers2019sentence} library, specifically employing the \texttt{all-MiniLM-L6-v2} model for sentence embedding.}

\item \textbf{Generating Adversarial passage and filtering:} We used the LLM to create supporting passages for the sentences generated in Step 2.
Similar to step 2, any adversarial passage containing the answer string was excluded.
\end{enumerate}
Throughout this process, we employed the OpenAI's \texttt{gpt-3.5-turbo-0125} model with default generation parameters as our LLM, and used SpaCy NER model\footnote{We used the \texttt{en\_core\_web\_trf} model. The link is as follows. \href{https://spacy.io/models/en}{https://spacy.io/models/en}}for named entity recognition. All the prompts we used can be found in Table \ref{tab:adv_prompts} in Appendix \ref{sec:prompts}.

Afterward, a single adversarial document is randomly inserted among the top-5 retrieved documents. This adversarial addition is semantically very similar to the question but is unrelated to the answer, thus acting as a distraction for the RALM.

\begin{table}[h]
\centering
\resizebox{\columnwidth}{!}{%
\begin{tabular}{l|l}
\hline
Question & Who got the first nobel prize in physics? \\ \hline
Answer & Wilhelm Conrad Röntgen \\ \hline
\begin{tabular}[c]{@{}l@{}}Answer\\ Sentence\end{tabular} & \begin{tabular}[c]{@{}l@{}}Wilhelm Conrad Röntgen was awarded the \\ first Nobel Prize in Physics.\end{tabular} \\ \hline
\begin{tabular}[c]{@{}l@{}}Adversarial\\ Sentence\end{tabular} & \begin{tabular}[c]{@{}l@{}} \textbf{Jesse Douglas} was the first recipient of the \textbf{Fields Medal} \\ \end{tabular} \\ \hline
\begin{tabular}[c]{@{}l@{}}Adversarial\\ Passage\end{tabular} & \begin{tabular}[c]{@{}l@{}}\textbf{Jesse Douglas}, an American mathematician, was awarded \\  \textbf{the first Fields Medal} in 1936 during the International Congress \\  of Mathematicians in Oslo. He was recognized for his work \\ on the Plateau problem, an important problem...  \end{tabular} \\ \hline
\end{tabular}%
}
\caption{An example of adversarial document generated by GenADV.}
\label{tab:genadv_ex}
\end{table}

\subsubsection{RAD score}
Our objective is to observe how the performance of the RALM changes when adversarial documents are added. Therefore, a mere decrease in the Exact Match (EM) score may not suffice for precise analysis. 
To systematically study this, first, we determine the \ARA (\textbf{ARA}), which consists of instances where the model provides correct answers with retrieved documents. 
We then define the \textbf{ARA-Add} as the instances in the ARA that maintained correctness even when an extra document was added.
Consequently, the \RAD\ score is calculated as follows:
\begin{equation}
    \textbf{RAD} = \frac{\text{\# of ARA-Add}}{\text{\# of ARA}} \times 100
\end{equation}
Using the RAD score, we can precisely analyze the impact that the addition of documents has on the results of the RALM.

\subsubsection{Results and Analysis}
\begin{figure*}
    \centering
    \includegraphics[width=1.0\textwidth]{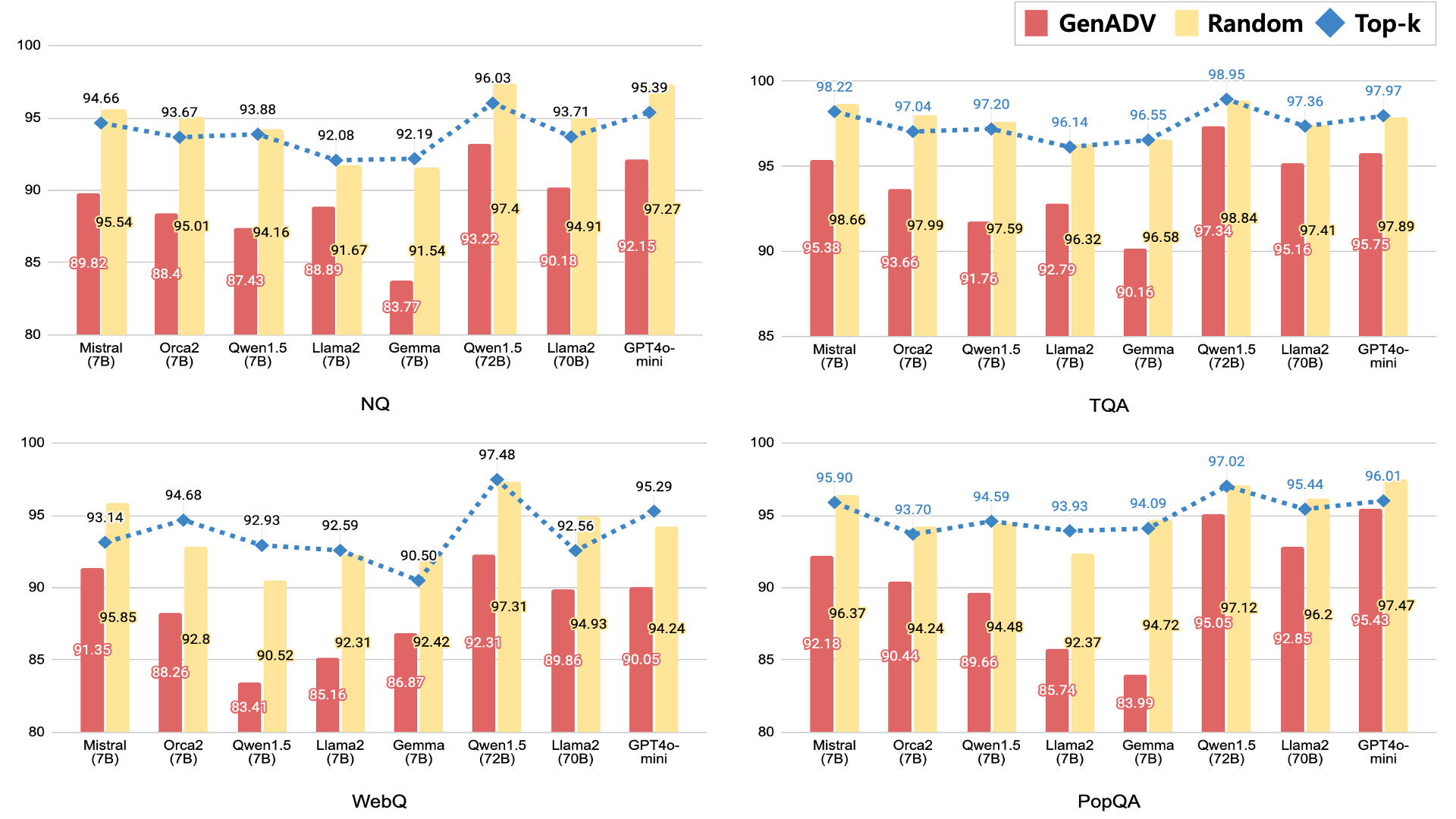}
    \caption{Experimental results on the effects of adding documents. The y-axis represents RAD score. GenADV refers to adding an adversarial document, Random to adding a randomly selected document, and Top-k to adding the sixth highest-ranked document.}
    \label{fig:adv_results}
\end{figure*}
To assess the effectiveness of GenADV, we compared the RAD scores based on the type of additional document used: a random document (selected randomly from the retrieved documents of other questions) and a top-k document (in the top-5 setting, this refers to the 6th highest-ranked document).


\textbf{RALMs are Not Robust to Adversaries} Figure \ref{fig:adv_results} shows the results of our experiments. In this experiment, we used the \textit{Normal prompt} to obtain the ARA and ARA-Add. 
Contrary to expectations, adding the top-k documents (referred to as Top-k in the figure) did not result in an RAD close to 100. In fact, in some cases, the RAD was lower than when random documents were added.
Particularly in the NQ, RAD was lower in five out of seven models.
This suggests that the retrieved documents can contain adversarial information that distracts the model, and that ignoring such documents can be more challenging for the RALM than disregarding completely unrelated documents. 
Moreover, the GenADV approach consistently resulted in the lowest RAD across all datasets and models, indicating that our method has the most significant distracting effect on the models. 

For example, with Gemma, in the NQ and PopQA datasets, RAD scores of 83.77 and 83.99 were reported, respectively.
This indicates that adding just one adversarial passage can induce hallucinations in about 17 out of 100 answers.
These results suggest that merely increasing the number of retrieved documents may have limited effects on enhancing the performance of open-domain QA, contrary to previous studies \citep{ram2023context} and that high-ranked documents also can distract RALMs.

\textbf{Challenges with Adversarial-Unanswerable Scenario}
We also created a scenario called \textit{adversarial unanswerable} scenario. 
This scenario involves a situation where the retrieved documents contain no correct answers (\textit{unanswerable}) while also including adversarial information (\textit{adversarial}). 
Consequently, RALMs should be able to identify unanswerability without being distracted by adversarial information.

\begin{table}[ht]
\resizebox{\columnwidth}{!}{%
\begin{tabular}{clcc|cc}
\hline
Dataset & \multicolumn{1}{c}{Models} & \begin{tabular}[c]{@{}c@{}}GenADV\\ (Ans.)\end{tabular} & \begin{tabular}[c]{@{}c@{}}Random\\ (Ans.)\end{tabular} & \begin{tabular}[c]{@{}c@{}}GenADV\\ (Unans.)\end{tabular} & \begin{tabular}[c]{@{}c@{}}Random\\ (Unans.)\end{tabular} \\ \hline
\multirow{7}{*}{NQ} & Mistral & \textbf{89.26} & 94.84 & \textbf{49.25} & 74.63 \\
 & Orca2 & \textbf{89.15} & 94.57 & \textbf{44.44} & 69.44 \\
 & Qwen1.5 & \textbf{86.35} & 92.73 & \textbf{64.86} & 89.86 \\
 & Gemma & \textbf{78.54} & 90.83 & \textbf{83.15} & 94.14 \\
 & Qwen1.5* & \textbf{91.45} & 96.76 & \textbf{69.41} & 88.24 \\
 & Llama2* & \textbf{91.24} & 94.88 & \textbf{68.42} & 85.09 \\ 
 & GPT4o-mini & \textbf{90.94} & 97.34 & \textbf{84.58} & 92.89 \\ \hline
\multirow{7}{*}{TQA} & Mistral & \textbf{95.38} & 98.73 & \textbf{48.61} & 70.14 \\
 & Orca2 & \textbf{93.09} & 97.45 & \textbf{45.24} & 77.38 \\
 & Qwen1.5 & \textbf{90.65} & 97.17 & \textbf{64.94} & 93.77 \\
 & Gemma & \textbf{86.49} & 95.18 & \textbf{78.46} & 89.37 \\
 & Qwen1.5* & \textbf{96.44} & 98.84 & \textbf{77.22} & 95.44 \\
 & Llama2* & \textbf{95.46} & 98.22 & \textbf{54.84} & 83.87 \\
 & GPT4o-mini & \textbf{93.52} & 97.98 & \textbf{81.78} & 94.13 \\ \hline
\end{tabular}%
}
\caption{Experimental results on the adversarial unanswerable scenario. Llama2 (7B) was excluded from the results because it correctly identified fewer than 30 unanswerable examples. 
* indicates largest models within the family (70B for Llama2 and 72B for Qwen1.5, respectively); all others are 7B models.}
\label{tab:adv_unans}
\end{table}

Typically, when retriever fails to fetch accurate information, there is a high likelihood of encountering only adversarial information that is related to the correct answer.
Therefore, this scenario is both realistic and crucial.
To test this scenario, we used the \textit{Unans Prompt} to obtain the ARA\footnote{Also we replaced the original answers with \textit{unanswerable} for unanswerable examples.}, then categorized the ARA into answerable and unanswerable examples. 
Subsequently, we identified the ARA-Add for each category and calculated the RAD score respectively. 
In this experiment, answerable examples refer to those that are not unanswerable. Through this experiment, we were able to assess the impact of document addition on the performance of RALMs for both answerable and unanswerable examples.
\begin{table*}[ht]
\centering
\renewcommand{\arraystretch}{0.9}
\begin{tabular}{lcccccccccccc}
\hline
 & \multicolumn{3}{c}{\textbf{NQ}} & \multicolumn{3}{c}{\textbf{TQA}} & \multicolumn{3}{c}{\textbf{WebQ}} & \multicolumn{3}{c}{\textbf{PopQA}} \\
\multicolumn{1}{c}{Models} & Acc & \begin{tabular}[c]{@{}c@{}}Acc\\ (C)\end{tabular} & \multicolumn{1}{c|}{\begin{tabular}[c]{@{}c@{}}Acc\\ (NC)\end{tabular}} & Acc & \begin{tabular}[c]{@{}c@{}}Acc\\ (C)\end{tabular} & \multicolumn{1}{c|}{\begin{tabular}[c]{@{}c@{}}Acc\\ (NC)\end{tabular}} & Acc & \begin{tabular}[c]{@{}c@{}}Acc\\ (C)\end{tabular} & \multicolumn{1}{c|}{\begin{tabular}[c]{@{}c@{}}Acc\\ (NC)\end{tabular}} & Acc & \begin{tabular}[c]{@{}c@{}}Acc\\ (C)\end{tabular} & \begin{tabular}[c]{@{}c@{}}Acc\\ (NC)\end{tabular} \\ \hline
\multicolumn{1}{l|}{Mistral} & 48.68 & 35.88 & \multicolumn{1}{c|}{66.63} & 65.58 & 30.09 & \multicolumn{1}{c|}{88.96} & 43.17 & 32.34 & \multicolumn{1}{c|}{62.18} & 59.77 & 52.42 & 80.84 \\
\multicolumn{1}{l|}{Orca2} & 25.84 & 0.35 & \multicolumn{1}{c|}{61.58} & 50.95 & 0.18 & \multicolumn{1}{c|}{84.4} & 21.21 & 0.48 & \multicolumn{1}{c|}{57.56} & 19.88 & 0.14 & 76.48 \\
\multicolumn{1}{l|}{Qwen1.5} & 23.45 & 9.65 & \multicolumn{1}{c|}{42.8} & 46.56 & 5.39 & \multicolumn{1}{c|}{73.69} & 24.64 & 17.13 & \multicolumn{1}{c|}{37.82} & 23.61 & 11.56 & 58.12 \\
\multicolumn{1}{l|}{Llama2} & 21.39 & 5.27 & \multicolumn{1}{c|}{43.97} & 46.33 & 0.24 & \multicolumn{1}{c|}{76.7} & 23.04 & 13.29 & \multicolumn{1}{c|}{40.13} & 16.62 & 0.51 & 62.81 \\
\multicolumn{1}{l|}{Gemma} & 40.5 & 34 & \multicolumn{1}{c|}{49.61} & 57.16 & 32.01 & \multicolumn{1}{c|}{73.73} & 34.71 & 31.38 & \multicolumn{1}{c|}{40.55} & 44.85 & 38.5 & 63.06 \\
\multicolumn{1}{l|}{Qwen1.5*} & 27.99 & 6.66 & \multicolumn{1}{c|}{57.88} & 53.17 & 3.53 & \multicolumn{1}{c|}{85.88} & 25.48 & 7.31 & \multicolumn{1}{c|}{57.35} & 26.67 & 12.66 & 66.81 \\
\multicolumn{1}{l|}{Llama2*} & 34.67 & 20.68 & \multicolumn{1}{c|}{54.28} & 52.44 & 6.3 & \multicolumn{1}{c|}{82.85} & 29.14 & 20.6 & \multicolumn{1}{c|}{44.12} & 39.59 & 27.13 & 75.3 \\
\multicolumn{1}{l|}{GPT4o-mini} & 26.65 & 11.52 & \multicolumn{1}{c|}{47.86} & 51.56 & 7.95 & \multicolumn{1}{c|}{80.29} & 21.36 & 10.54 & \multicolumn{1}{c|}{40.34} & 30.65 & 21.15 & 57.88 \\ \hline
\end{tabular}
\caption{Experimental results on identifying conflicts. Acc (C) means the accuracy for conflicting examples, Acc (NC) means the accuracy for non-conflicting examples, and Acc refers to the accuracy for all answerable examples. * indicates largest models within the family (70B for Llama2 and 72B for Qwen1.5, respectively); all others are 7B models.}
\label{tab:conflict_results}
\end{table*}
Table \ref{tab:adv_unans} shows the experimental results for the NQ and TQA datasets under adversarial-unanswerable scenarios, with random method added for comparison. 
The numbers in the table represent the RAD scores.
Across all models, the impact of the adversary was more pronounced on unanswerable examples (Unans.) compared to answerable ones (Ans.). 
While Gemma showed relative robustness to unanswerable scenarios in NQ, it displayed the lowest RAD scores for answerable examples. 
These findings indicate that LLMs may struggle more with identifying examples as unanswerable when they fail to retrieve correct answers, particularly in adversarial settings.
This result shows that it is more challenging for the RALM to confirm the absence of an answer where none exists than to find the correct answer where one is present. 
Thus, the close relationship between unanswerable examples and adversaries in real-world contexts implies that merely supplying gold documents with the correct answers is inadequate to address all challenges.

\subsection{Identifying Conflicting Information}
In this experiment, we evaluate the robustness of LLMs based on two criteria. First, we evaluate the model's ability to detect conflicts in the retrieved documents in a zero-shot setting (conflict detection). Second, we assess whether the model can generate accurate responses when presented with conflicting documents (stubbornness).

\textbf{Conflict Detection} For this evaluation, we used the \textit{Conflict prompt} specifically designed for conflict detection and measured how accurately the model identified the presence of a \textit{conflict}. This experiment was conducted on the answerable examples, with accuracy as the metric.

\textbf{Stubbornness} To evaluate this, we used a \textit{Normal prompt} and measured how well the model generated original answers to the questions despite conflicting information within the documents.
This experiment, evaluating stubbornness in retrieval augmentation results, differs from previous studies focused on stubbornness in parametric knowledge \citep{mallen2023not,xie2023adaptive}.
We conducted this experiment on the ARA, using accuracy as our metric.


\subsubsection{Crafting conflicting documents}
\label{subsubsec:craft_conflict}
We followed \citet{xie2023adaptive}'s method to create conflicting documents. 
Specifically, similar to the process of creating adversarial documents, we first generated an answer sentence. 
Unlike \citep{xie2023adaptive}, we do not perform random substitutions within the same type entity; instead, replaced answer entities in the answer sentence with similar entities of the same type to create a conflicting sentence. 
This is because we assume a retrieval scenario, and therefore, the conflicting information must also contain information similar to the original answer. 
We utilized the SpaCy\footnote{We used the \texttt{en\_core\_web\_lg} model.} token embedding model to calculate the cosine similarity between entities, and to exclude aliases, we substituted entities with a cosine similarity score of 0.8 or lower.\footnote{The entity pool for substitutions was created by extracting entities from all texts in the \texttt{Wikitext-103-raw-v1} dataset} 
Then, using an LLM, we generated a supporting conflicting passage and a single conflict passage was randomly inserted among the top-5 documents, similar to our approach with adversarial documents.
\subsubsection{Results and Analysis}

\textbf{LLMs as Poor Conflict Detectors} Table \ref{tab:conflict_results} shows the experimental results for identifying conflicting information. Across all datasets and models, the accuracy for conflicting examples was notably low. 
Although Mistral and Gemma showed relatively high accuracy, even the largest models of Qwen1.5 and Llama2 performed worse than these models. 
Despite the fact that we carefully crafted the conflicting information to be highly intuitive, the results indicate that LLMs significantly struggle to detect conflicts within documents. 
This underscores the challenge LLMs face in recognizing conflicts, particularly when they are involved in retrieving and generating content from multiple sources.

\textbf{RALMs are Vulnerable to Misinformation} Table \ref{tab:conflict_stubborn} illustrates the results of experiments testing the stubbornness of RALMs. 
We analyzed the proportion of examples in the ARA that either maintained correctness (A$\rightarrow$A), sourced answers from conflicting documents (A$\rightarrow$C), or did neither (A$\rightarrow$U) when the conflicting document was added.
The results highlight the susceptibility of LLMs to misinformation.
Gemma retained only 57.88\% accuracy in the NQ dataset, while even the highest-performing Llama2 managed only 75.92\%. This suggests that in the presence of deliberate misinformation, LLMs are prone to abandoning correct answers in favor of the erroneous information.
Particularly in the real world, if contaminated information such as fake news from the internet is retrieved, it implies that RALMs can potentially provide incorrect responses based on such sources.
\begin{table}[H]
\centering
\resizebox{\columnwidth}{!}{%
\begin{tabular}{lccc|ccc}
 & \multicolumn{3}{c|}{\textbf{NQ}} & \multicolumn{3}{c}{\textbf{TQA}} \\
\multicolumn{1}{c}{Models} & A$\rightarrow$A & A$\rightarrow$C & A$\rightarrow$U & A$\rightarrow$A & A$\rightarrow$C & A$\rightarrow$U \\ \hline
\multicolumn{1}{l|}{Mistral} & 68.34 & 25.61 & 6.05 & 79.74 & 16.84 & 3.42 \\
\multicolumn{1}{l|}{Orca2} & 69.36 & 23.63 & 7.01 & 76.49 & 18.13 & 5.38 \\
\multicolumn{1}{l|}{Qwen1.5} & 66.45 & 24.36 & 9.19 & 68.83 & 23.34 & 7.83 \\
\multicolumn{1}{l|}{Llama2} & 75.92 & 13.23 & 10.88 & 80.66 & 11.07 & 8.27 \\
\multicolumn{1}{l|}{Gemma} & 57.88 & 33.45 & 8.67 & 63.11 & 28.55 & 8.35 \\ 
\multicolumn{1}{l|}{Qwen1.5*} & 82.64 & 12.37 & 4.99 & 90.5 & 7.07 & 2.43 \\
\multicolumn{1}{l|}{Llama2*} & 74.02 & 18.56 & 7.42 & 82.23 & 12.52 & 5.25 \\
\multicolumn{1}{l|}{GPT4o-mini} & 74.84 & 16.14 & 9.02 & 82.07 & 11.04 & 6.89 \\\hline
\end{tabular}%
}
\caption{Experimental results on the changes in answers when conflicting documents are added. A$\rightarrow$A (Answer to Answer) indicates cases where the answer remained the same after the addition of a conflicting document, A$\rightarrow$C (Answer to Conflict) indicates cases where the answer was based on information in the conflicting document, and A$\rightarrow$U (Answer to Uncertain) refers to all other cases.}
\label{tab:conflict_stubborn}
\end{table}
\section{Conclusion}
In this study, we conducted a comprehensive evaluation of the robustness of RALMs under various imperfect retrieval conditions. Our findings revealed significant challenges faced by these models in handling unanswerablity, adversarial and conflicting information.

Through extensive experiments, we demonstrated that RALMs struggle to identify unanswerable scenarios, often hallucinating responses even when the retrieved documents do not contain the answer. 
Additionally, we introduced a new method, GenADV, for generating adversarial information, which proved highly effective in distracting the models and causing them to abandon correct answers.
Furthermore, our results highlighted the vulnerability of RALMs to conflicting information, as they exhibited poor performance in both detecting conflicts within the retrieved documents and generating accurate responses in the presence of such conflicts. Our study provides a foundation for evaluating the robustness of RALMs, crucial for their safe use. Based on this foundation, further exploration into developing robust models is necessary.

\section*{Acknowledgements}
This work was supported in part by the National Research Foundation of Korea (NRF) grant (RS-2023-00280883, RS-2023-00222663), by the National Super computing Center with super computing resources including technical support (KSC-2023-CRE-0176), and partially supported by New Faculty Startup Fund from Seoul National University

\bibliography{custom}
\bibliographystyle{acl_natbib}

\iftaclpubformat

\onecolumn






  
\appendix
\section{Prompts}
\label{sec:prompts}
To facilitate the reproducibility of our experiments, we are releasing all the prompts used in our study. Table \ref{tab:prompts} and \ref {tab:adv_prompts} show the instructions. The curly brackets denote placeholders where actual values will be inserted.

\begin{table}[h]
\centering
\scriptsize
\renewcommand{\arraystretch}{0.9}
\begin{tabular}{c|l}
\hline
\textbf{Name} & \multicolumn{1}{c}{\textbf{Instruction}} \\ \hline
\textbf{Normal} & \begin{tabular}[c]{@{}l@{}}Documents:\{retrieved documents\}\\ Use the above documents to answer the subsequent question. Please provide the answer as a single word \\or term, without forming a complete sentence.\\ Question: \{question\}\\ Answer:\end{tabular} \\ \hline
\textbf{Unans} & \begin{tabular}[c]{@{}l@{}}Documents:\{retrieved documents\}\\ Use the above documents to answer the subsequent question. Please provide the answer as a single word \\or term, without forming a complete sentence. If the answer cannot be found, write 'unanswerable'\\ Question: \{question\}\\ Answer:\end{tabular} \\ \hline
\textbf{Conflict} & \begin{tabular}[c]{@{}l@{}}Documents:\{retrieved documents\}\\ Use the above documents to answer the subsequent question. Please provide the answer as a single word \\or term, without forming a complete sentence. If multiple documents present different answers, please \\respond with 'conflict' to indicate\\ the presence of conflicting information.\\ Question: \{question\}\\ Answer:\end{tabular} \\ \hline
\end{tabular}
\caption{The prompts used in the experiment.}
\label{tab:prompts}
\end{table}

\begin{table}[h]
\centering
\scriptsize
\renewcommand{\arraystretch}{0.9}
\begin{tabular}{c|l}
\hline
\textbf{Step} & \multicolumn{1}{c}{\textbf{Instruction}} \\ \hline
\textbf{\begin{tabular}[c]{@{}c@{}}Answer\\ Sentence\\ Generation\end{tabular}} & \begin{tabular}[c]{@{}l@{}}Please write a single sentence using the following question and answer. The sentence should include \\ the answer and be as realistic as possible.:\\ Question: \{question\}\\ Answer: \{answer\}\\ Sentence:\end{tabular} \\ \hline
\textbf{\begin{tabular}[c]{@{}c@{}}Adversarial \\ Sentence\\ Generation\end{tabular}} & \begin{tabular}[c]{@{}l@{}}Rewrite the sentence by replacing the specified words with others, ensuring that the new sentence \\ retains a meaning as close as possible to the original while not being identical. The words to replace \\are named entities, which should be substituted with entities of the same type. The revised sentence \\must also remain factually accurate.\\ Original sentence: \{answer sentence\}\\ Words to replace: \{named entities\}\\ Revised sentence:\end{tabular} \\ \hline
\textbf{\begin{tabular}[c]{@{}c@{}}Adversarial\\ Passage\\ Generation\end{tabular}} & \begin{tabular}[c]{@{}l@{}}Given a claim, write a concise, factual passage using 50 to 100 words to support it. Please write the \\passage in the style of Wikipedia:\\ Claim: \{adversarial sentence\}\\ Passage:\end{tabular} \\ \hline
\end{tabular}%
\caption{The prompts used for crafting adversarial information.}
\label{tab:adv_prompts}
\end{table}

\section{Baseline performance}
\label{sec:baseline}

We evaluated the performance of models on closed QA as well as retrieval augmented QA. 
These experiments followed the settings outlined in \sref{sec:setup}.
Closed QA refers to the task where no retrieved documents are provided, allowing us to gauge how much knowledge the model possesses about the dataset, known as \textit{parametric knowledge}. 
For closed QA, we used the following prompt: \textit{Answer the following question. Please provide the answer as a single word or term, without forming a complete sentence. Q: \{question\} A:}

To comprehensively assess QA performance, we additionally calculated the exact match score (EM) and F1 score (F1), following the \citep{izacard2022few}. The results of these experiments are presented in Table \ref{tab:clsoed_qa_baseline}.

Next, for performance in retrieval augmentation, we provided the top-5 retrieved documents and used a \textit{Normal prompt} for inference. We also computed additional metrics. The results are shown in Table \ref{tab:rag_baseline}.

Finally, to further investigate the model's parametric knowledge, we calculated the accuracy rate of correct answers for examples where the top-5 retrieved documents did not contain the answer (same as "unanswerable" in \sref{subsec:unans_setting}). 
If the model correctly answered the question without any relevant information provided, it likely relied on its parametric knowledge. 
Therefore, a higher rate suggests greater parametric knowledge. 
We define this metric as the Parametric Answer Rate (PAR). The results for PAR are presented in Table \ref{tab:parametric_answer}.

\begin{table*}[ht]
\centering
\renewcommand{\arraystretch}{0.85}
\begin{tabular}{lcccccccccccc} \Xhline{1.5\arrayrulewidth}
\hline
\multicolumn{13}{c}{\textbf{Baselines without retrieval}} \\ \cline{2-13} 
\multicolumn{1}{c}{} & \multicolumn{3}{c|}{\textbf{NQ}} & \multicolumn{3}{c|}{\textbf{TQA}} & \multicolumn{3}{c|}{\textbf{WebQ}} & \multicolumn{3}{c}{\textbf{PopQA}} \\ \cline{2-13} 
\multicolumn{1}{c}{Models} & Acc & EM & \multicolumn{1}{c|}{F1} & Acc & EM & \multicolumn{1}{c|}{F1} & Acc & EM & \multicolumn{1}{c|}{F1} & Acc & EM & F1 \\ \hline
\multicolumn{1}{l|}{Mistral} & 33.55 & 3.38 & \multicolumn{1}{c|}{12.92} & 60.44 & 25.29 & \multicolumn{1}{c|}{37.63} & 43.90 & 4.63 & \multicolumn{1}{c|}{21.35} & 24.52 & 6.09 & 13.27 \\
\multicolumn{1}{l|}{Orca2} & 32.63 & 4.02 & \multicolumn{1}{c|}{16.93} & 55.74 & 21.77 & \multicolumn{1}{c|}{38.17} & 43.95 & 6.99 & \multicolumn{1}{c|}{25.73} & 24.84 & 5.04 & 15.52 \\
\multicolumn{1}{l|}{Qwen1.5} & 16.29 & 12.30 & \multicolumn{1}{c|}{18.86} & 34.88 & 30.06 & \multicolumn{1}{c|}{37.08} & 24.90 & 13.39 & \multicolumn{1}{c|}{28.44} & 15.36 & 13.02 & 17.23 \\
\multicolumn{1}{l|}{Llama2} & 14.96 & 12.02 & \multicolumn{1}{c|}{21.58} & 38.50 & 37.34 & \multicolumn{1}{c|}{47.26} & 19.09 & 15.75 & \multicolumn{1}{c|}{32.03} & 16.44 & 16.14 & 21.85 \\
\multicolumn{1}{l|}{Gemma} & 14.96 & 3.99 & \multicolumn{1}{c|}{11.49} & 37.64 & 14.55 & \multicolumn{1}{c|}{27.22} & 23.47 & 5.02 & \multicolumn{1}{c|}{19.19} & 15.06 & 3.40 & 9.31 \\
\multicolumn{1}{l|}{Qwen1.5*} & 35.41 & 23.80 & \multicolumn{1}{c|}{34.95} & 64.08 & 56.02 & \multicolumn{1}{c|}{64.93} & 43.21 & 19.34 & \multicolumn{1}{c|}{37.69} & 31.73 & 22.89 & 28.95 \\
\multicolumn{1}{l|}{Llama2*} & 24.82 & 20.36 & \multicolumn{1}{c|}{31.16} & 55.26 & 54.12 & \multicolumn{1}{c|}{64.39} & 23.62 & 20.28 & \multicolumn{1}{c|}{36.23} & 25.53 & 25.35 & 29.69 \\
\multicolumn{1}{l|}{GPT4o-mini} & 29.73 & 29.70 & \multicolumn{1}{c|}{41.27} & 58.94 & 59.44 & \multicolumn{1}{c|}{69.84} & 24.26 & 22.74 & \multicolumn{1}{c|}{38.69} & 27.58 & 27.38 & 32.90 \\ \hline
\Xhline{1.5\arrayrulewidth}
\end{tabular}

\caption{Experimental Results for Closed QA. * indicates largest models within the family (70B for Llama2 and 72B for Qwen1.5, respectively)}
\label{tab:clsoed_qa_baseline}
\end{table*}

\begin{table*}[ht]
\renewcommand{\arraystretch}{0.85}
\begin{tabular}{lcccccccccccc}\Xhline{1.5\arrayrulewidth}
\hline
\multicolumn{13}{c}{\textbf{Baselines with retrieval}} \\ \cline{2-13} 
\multicolumn{1}{c}{} & \multicolumn{3}{c|}{\textbf{NQ}} & \multicolumn{3}{c|}{\textbf{TQA}} & \multicolumn{3}{c|}{\textbf{WebQ}} & \multicolumn{3}{c}{\textbf{PopQA}} \\ \cline{2-13} 
\multicolumn{1}{c}{Models} & Acc & EM & \multicolumn{1}{c|}{F1} & Acc & EM & \multicolumn{1}{c|}{F1} & Acc & EM & \multicolumn{1}{c|}{F1} & Acc & EM & F1 \\ \hline
\multicolumn{1}{l|}{Mistral} & 51.61 & 12.19 & \multicolumn{1}{c|}{26.95} & 70.67 & 46.73 & \multicolumn{1}{c|}{59.12} & 47.05 & 9.30 & \multicolumn{1}{c|}{25.60} & 57.12 & 21.55 & 33.21 \\
\multicolumn{1}{l|}{Orca2} & 49.34 & 5.57 & \multicolumn{1}{c|}{22.15} & 69.11 & 17.05 & \multicolumn{1}{c|}{37.11} & 44.39 & 4.08 & \multicolumn{1}{c|}{22.87} & 53.14 & 2.34 & 23.40 \\
\multicolumn{1}{l|}{Qwen1.5} & 39.36 & 33.63 & \multicolumn{1}{c|}{43.57} & 63.40 & 59.91 & \multicolumn{1}{c|}{67.43} & 33.86 & 19.73 & \multicolumn{1}{c|}{35.60} & 47.23 & 42.15 & 47.47 \\
\multicolumn{1}{l|}{Llama2} & 34.90 & 32.58 & \multicolumn{1}{c|}{42.54} & 60.05 & 59.11 & \multicolumn{1}{c|}{66.95} & 27.61 & 18.31 & \multicolumn{1}{c|}{33.12} & 40.30 & 39.69 & 44.42 \\
\multicolumn{1}{l|}{Gemma} & 39.81 & 26.32 & \multicolumn{1}{c|}{35.94} & 60.69 & 48.59 & \multicolumn{1}{c|}{58.35} & 31.89 & 15.50 & \multicolumn{1}{c|}{29.74} & 45.57 & 32.92 & 39.27 \\
\multicolumn{1}{l|}{Qwen1.5*} & 48.89 & 39.39 & \multicolumn{1}{c|}{50.36} & 71.49 & 65.31 & \multicolumn{1}{c|}{73.78} & 42.27 & 22.93 & \multicolumn{1}{c|}{39.58} & 51.78 & 43.89 & 50.42 \\
\multicolumn{1}{l|}{Llama2*} & 40.00 & 35.68 & \multicolumn{1}{c|}{47.14} & 63.80 & 63.17 & \multicolumn{1}{c|}{71.52} & 32.04 & 20.18 & \multicolumn{1}{c|}{35.36} & 48.76 & 45.22 & 49.94 \\
\multicolumn{1}{l|}{GPT4o-mini} & 41.16 & 39.97 & \multicolumn{1}{c|}{51.11} & 65.53 & 65.05 & \multicolumn{1}{c|}{74.73} & 30.81 & 25.10 & \multicolumn{1}{c|}{40.22} & 48.85 & 48.05 & 52.67 \\ \hline
\Xhline{1.5\arrayrulewidth}
\end{tabular}%
\caption{Experimental results for retrieval augmentation QA. * indicates largest models within the family (70B for Llama2 and 72B for Qwen1.5, respectively)}
\label{tab:rag_baseline}
\end{table*}

\begin{table*}[ht]
\tiny
\setlength{\tabcolsep}{30pt}
\renewcommand{\arraystretch}{0.85}
\begin{tabular}{lcccc}\Xhline{1.5\arrayrulewidth}
\hline
\multicolumn{5}{c}{Parametric Answer Rate} \\ \hline
\multicolumn{1}{c|}{Models} & NQ & TQA & WebQ & PopQA \\ \hline
\multicolumn{1}{l|}{Mistral} & 4.82 & 11.62 & 5.96 & 2.83 \\
\multicolumn{1}{l|}{Orca2} & 6.13 & 12.21 & 6.13 & 2.79 \\
\multicolumn{1}{l|}{Qwen1.5} & 1.49 & 5.25 & 1.82 & 1.62 \\
\multicolumn{1}{l|}{Llama2} & 1.41 & 5.75 & 1.39 & 1.32 \\
\multicolumn{1}{l|}{Gemma} & 0.26 & 2.97 & 1.66 & 1.38 \\
\multicolumn{1}{l|}{Qwen1.5*} & 5.61 & 15.67 & 4.58 & 2.55 \\
\multicolumn{1}{l|}{Llama2*} & 3.24 & 10.46 & 3.05 & 2.02 \\
\multicolumn{1}{l|}{GPT4o-mini} & 4.47 & 14.32 & 2.64 & 1.94 \\ \hline
\Xhline{1.5\arrayrulewidth}\end{tabular}
\caption{Results of parametric answer rate. We used the Normal prompt for inference. A higher rate indicates that the model correctly answered more questions even without the relevant document being provided, suggesting a greater amount of parametric knowledge about the dataset.} 
\label{tab:parametric_answer}
\end{table*}

\section{Human evaluation}
\label{sec:human_eval}

\subsection{GenADV}
\label{subsec:genadv_eval}

\begin{table}[htbp]
\centering
\renewcommand{\arraystretch}{0.9}
\begin{tiny}
\begin{tabular}{cl}
\hline
\multicolumn{1}{c|}{\textbf{Score}} & \multicolumn{1}{c}{\textbf{Description}} \\ \hline
\multicolumn{2}{c}{\textbf{Consistency}} \\ \hline
\multicolumn{1}{c|}{1} & \begin{tabular}[c]{@{}l@{}}The passage is very awkward, with poor sentence flow.\end{tabular} \\ \hline
\multicolumn{1}{c|}{2} & \begin{tabular}[c]{@{}l@{}}The passage is somewhat natural but has minor issues in expression or flow.\end{tabular} \\ \hline
\multicolumn{1}{c|}{3} & \begin{tabular}[c]{@{}l@{}}The passage is very natural and flows smoothly.\end{tabular} \\ \hline
\multicolumn{2}{c}{\textbf{Similarity}} \\ \hline
\multicolumn{1}{c|}{1} & \begin{tabular}[c]{@{}l@{}}The passage is completely unrelated to the question's topic.\end{tabular} \\ \hline
\multicolumn{1}{c|}{2} & \begin{tabular}[c]{@{}l@{}}Some content or words in the passage are related to the question's topic.\end{tabular} \\ \hline
\multicolumn{1}{c|}{3} & \begin{tabular}[c]{@{}l@{}}Most of the passage content is closely related to the question's topic.\end{tabular} \\ \hline
\multicolumn{2}{c}{\textbf{Relevance}} \\ \hline
\multicolumn{1}{c|}{1} & \begin{tabular}[c]{@{}l@{}}It is impossible to infer the correct answer from the passage.\end{tabular} \\ \hline
\multicolumn{1}{c|}{2} & \begin{tabular}[c]{@{}l@{}}The passage content is related to the correct answer but does not directly provide it.\end{tabular} \\ \hline
\multicolumn{1}{c|}{3} & \begin{tabular}[c]{@{}l@{}}The passage content allows for direct or indirect inference of the correct answer.\end{tabular} \\ \hline
\end{tabular}
\end{tiny}
\caption{Guidelines provided to evaluators for assessing passages generated by GenADV}
\label{tab:guide}
\end{table}

To validate the effectiveness of our GenADV, we conducted a human evaluation. We randomly sampled 25 questions, answers, and generated adversarial passages from NQ, TQA, WebQ, and PopQA datasets. 
These samples were then evaluated on three criteria: \textit{consistency, similarity}, and \textit{relevance}, each scored on a scale from 1 to 3.
Consistency assesses the fluency and coherence of the passage. Similarity measures how closely the passage's topic aligns with the question's topic. 
Relevance evaluates how well the passage allows one to infer the answers. 
According to these criteria, a good adversarial passage should score high in consistency and similarity, but low in relevance. 
The guidelines provided to the evaluators can be found in Table \ref{tab:guide}. To ensure objectivity, we did not disclose the purpose of the evaluation to the evaluators, nor did we inform them that the passages were generated by AI.

\begin{table}[ht]
\renewcommand{\arraystretch}{0.85}
\setlength{\tabcolsep}{30pt}
\centering
\begin{tabular}{l|l}
\hline
\multicolumn{1}{c|}{\textbf{Criteria}} & \multicolumn{1}{c}{\textbf{Score}} \\ \hline
Consistency & 2.83 \\
Similarity & 2.24 \\
Relevance & 1.14 \\ \hline
\end{tabular}
\caption{Evaluation results for each criterion of passages generated by GenADV}
\label{tab:eval_result}
\end{table}

We selected 10 non-native English speakers with high proficiency in English as evaluators. Each of the 100 samples was reviewed by two evaluators, and the final score was the average of their individual scores. The average scores from the evaluation are shown in Table \ref{tab:eval_result}. These results indicate that while GenADV effectively generates passages related to the question's topic, the relevance to the actual answers remains low.

\subsection{Conflict}
We also manually verify the presence of conflicting information in the conflict passages we created. 
Similar to \ref{subsec:genadv_eval}, we sampled 25 instances from each dataset and evaluated them.
If the passage contained information that contradicted the original answer, we labeled it as a "conflict". 
If there was no contradiction, it was labeled as "non-conflict". 
Instances where the conflict was uncertain, such as when the passage was not relevant to the question or did not provide a clear answer, were labeled as "uncertain".
In our evaluation, 83 passages contained conflicting information, while only one passage had none, meaning that we successfully generated conflicting passages in over 80\% of the instances.

\end{document}
